\documentclass{article}

\usepackage[final]{neurips_2025}

\usepackage[utf8]{inputenc} 
\usepackage[T1]{fontenc}    
\usepackage{hyperref}       
\usepackage{url}            
\usepackage{booktabs}       
\usepackage{amsfonts}       
\usepackage{nicefrac}       
\usepackage{microtype}      
\usepackage{xcolor}         
\usepackage{amsmath,amssymb,amsfonts}
\usepackage{graphicx}
\usepackage{textcomp}
\usepackage{xcolor}
\def\BibTeX{{\rm B\kern-.05em{\sc i\kern-.025em b}\kern-.08em
    T\kern-.1667em\lower.7ex\hbox{E}\kern-.125emX}}

\usepackage{booktabs} 
\usepackage{import}
\usepackage{graphicx}
\usepackage{epstopdf}
\usepackage{amsmath}
\usepackage{subfigure}
\usepackage{diagbox}
\usepackage{float}
\graphicspath{{Fig/}}
\usepackage{booktabs} 
\usepackage{algorithm}
\usepackage{algorithmicx}
\usepackage{algpseudocode}
\usepackage{paralist}
\usepackage{multirow}
\usepackage{graphicx}
\usepackage{subfigure}
\usepackage{color}
\usepackage{hhline}
\usepackage{xcolor}
\usepackage{color,soul}
\usepackage{array}
\usepackage{diagbox}
\usepackage{hyperref}
\usepackage{verbatim} 
\usepackage{colortbl}
\usepackage{upgreek}

\usepackage{enumitem}
\usepackage{gensymb} 
\usepackage{ulem} 
\usepackage{amsmath,bm}
\usepackage{amssymb}
\usepackage{pifont}
\usepackage{filecontents}
\algdef{SE}[DOWHILE]{Do}{doWhile}{\algorithmicdo}[1]{\algorithmicwhile\ #1}%
\usepackage{amsmath, amssymb, amsthm}
\usepackage{algorithm}

\definecolor{deep_pink}{HTML}{D97692}

\title{AI's Euclid's Elements Moment: \\ From Language Models to Computable Thought}

\author{%
  Xinmin Fang  \\
  Department of Computer Science and Engineering\\
  University of Colorado Denver (CU Denver)\\
  Denver, CO 80204 \\
  \texttt{xinmin.fang@ucdenver.edu} \\
  \And
  Lingfeng Tao \\
  Department of Robotics and Mechatronics \\
  Kennesaw State University (KSU)\\
  Atlanta, GA, 30144 \\
  \texttt{ltao2@kennesaw.edu} \\
  \AND
  Zhengxiong Li \thanks{Disclaimer: This working paper is intended solely for academic discussion and exploratory analysis. It does not constitute investment advice or serve as a basis for any financial decision-making. All data and examples referenced are derived from publicly available media sources and industry reports. \\
  * This work is partially supported by US NSF Awards \#2426469 and \#2426470.
  } \\
  Department of Computer Science and Engineering \\
  University of Colorado Denver (CU Denver)\\
  Denver, CO 80204 \\
  \texttt{zhengxiong.li@ucdenver.edu} \\
}

\begin{document}

\maketitle
\begin{abstract}
This paper presents a comprehensive five-stage evolutionary framework for understanding the development of artificial intelligence, arguing that its trajectory mirrors the historical progression of human cognitive technologies. We posit that AI is advancing through distinct epochs, each defined by a revolutionary shift in its capacity for representation and reasoning, analogous to the inventions of cuneiform, the alphabet, grammar and logic, mathematical calculus, and formal logical systems. This "Geometry of Cognition" framework moves beyond mere metaphor to provide a systematic, cross-disciplinary model that not only explains AI's past architectural shifts—from expert systems to Transformers—but also charts a concrete and prescriptive path forward. Crucially, we demonstrate that this evolution is not merely linear but reflexive: as AI advances through these stages, the tools and insights it develops create a feedback loop that fundamentally reshapes its own underlying architecture. We are currently transitioning into a "Metalinguistic Moment," characterized by the emergence of self-reflective capabilities like Chain-of-Thought prompting and Constitutional AI. The subsequent stages, the "Mathematical Symbolism Moment" and the "Formal Logic System Moment," will be defined by the development of a computable calculus of thought, likely through neuro-symbolic architectures and program synthesis, culminating in provably aligned and reliable AI that reconstructs its own foundational representations. This work serves as the methodological capstone to our trilogy, which previously explored the economic drivers ("why") and cognitive nature ("what") of AI. Here, we address the "how," providing a theoretical foundation for future research and offering concrete, actionable strategies for startups and developers aiming to build the next generation of intelligent systems.

\end{abstract}

\section{Introduction: From Cognitive Augmentation to Computable Thought}

The rapid ascent of Artificial Intelligence (AI) represents a watershed moment in technological history, a revolution not of physical power but of cognitive capacity. This paper is the third and final installment of a trilogy dedicated to providing a multifaceted understanding of this transformation. Our first work, "Anchoring AI Capabilities in Market Valuations: The Capability Realization Rate Model and Valuation Misalignment Risk," explored the economic underpinnings of the current AI boom, the "why", by analyzing the intricate relationship between latent AI capabilities and their realized market value \cite{fang2025anchoring}. Our second paper, "Closer to Language than Steam: AI as the Cognitive Engine of a New Productivity Revolution," framed AI's fundamental nature—the "what"—as a cognitive engine analogous to the invention of written language, a tool that fundamentally augments and reshapes human intellect rather than merely automating physical labor \cite{tao2025closer}.

This paper now addresses the crucial question of "how": it proposes a developmental roadmap, a Geometry of Cognition, that charts the necessary evolutionary stages for AI to progress from its current, powerful linguistic capabilities toward a truly formal, computable, and ultimately general intelligence. We argue that the evolution of AI is not a random walk but a structured progression that closely parallels the historical development of humanity's own cognitive technologies. This is not merely a convenient metaphor; it is a deep structural parallel that offers both descriptive and prescriptive power. Just as human civilization advanced its cognitive reach through a series of foundational inventions—from concrete pictographs to abstract alphabets, from implicit language use to explicit grammar and logic, and from descriptive reasoning to the computable symbolism of calculus—AI is undergoing an analogous journey. Crucially, this evolution exhibits a reflexive quality: as AI develops more sophisticated cognitive tools, these tools recursively reshape the fundamental architecture of intelligence itself, creating a feedback loop that accelerates and transforms the evolutionary process.

Our central thesis is that AI's development can be understood through five distinct "moments," each representing a geometric leap in abstraction, structure, and self-reflection. The first is the \textbf{Cuneiform Moment}, where early AI, particularly expert systems, functioned as systems of record for narrow domains, much like cuneiform was used for accounting and administration. Following this was the \textbf{Alphabet Moment}, marked by the advent of the Transformer architecture, which provided a universal, abstract representational layer—a set of primitives (tokens and attention) that, like the alphabet, could encode any form of information, leading to the current explosion in generative capabilities. We are currently in the \textbf{Metalinguistic Moment}, where techniques like Chain-of-Thought (CoT) and Constitutional AI (CAI) signal the beginning of AI's self-reflection—its ability to analyze and structure its own thought processes, akin to the birth of grammar and logic. The next frontier is the \textbf{Mathematical Symbolism Moment}, where AI's reasoning will be formalized into a computable "calculus of thought," likely through neuro-symbolic systems and program synthesis, transforming reasoning from a linguistic act into a verifiable operation. The ultimate goal is the \textbf{Formal Logic System Moment}, where AI operates as a fully formalized system, whose safety, alignment, and behavior are provably correct, creating an artificial cognitive system parallel to human thought.

\begin{figure}[h!]
  \centering
  \includegraphics[width=\textwidth]{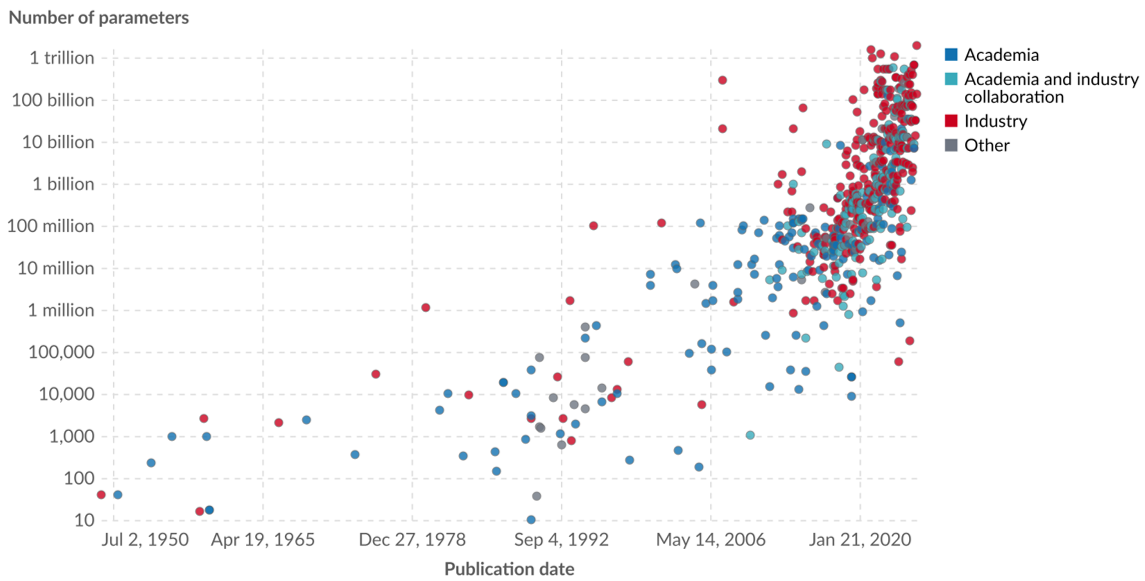}
  \caption{The exponential growth in the number of parameters in notable AI models. The dramatic inflection point around 2017-2018, coinciding with the introduction of the Transformer architecture, visually represents the "Alphabet Moment." This leap in scale underscores the shift from earlier, smaller-scale paradigms to the era of massive, generative models. Source: Our World in Data.}
  \label{fig:params}
\end{figure}

While individual historical analogies for AI have been proposed, this paper presents the first systematic, multi-stage framework that directly maps specific AI architectural paradigm shifts to their corresponding revolutions in human cognitive technology. This framework is not merely descriptive; it is prescriptive. It suggests that the challenges AI faces at each stage are homologous to those humanity encountered, and the solutions—greater abstraction, explicit rule systems, formal verification—represent a necessary, if not inevitable, path. For example, the transition from the brittleness of expert systems to the generality of Large Language Models (LLMs) was not just a matter of more data or faster computers; it required a fundamental solution to the problem of representation (see Figure~\ref{fig:params}). The historical record thus suggests that AI had to undergo an "Alphabet Moment" to transcend the limitations of its "Cuneiform Moment."

This paper will proceed by examining each of these five moments in detail. We will analyze the historical precedent, map it to the corresponding AI paradigm, and distill the core principles that govern each evolutionary stage. We then explore how this evolutionary process becomes self-modifying, with each stage's innovations recursively transforming the foundational architecture of AI itself. Finally, we will translate this theoretical framework into a practical roadmap, outlining concrete and actionable opportunities for startups and research labs seeking to build the future of AI. By understanding the geometry of AI's evolution, we can not only make sense of its past but also more deliberately and effectively construct its future.

\section{The Cuneiform Moment: AI as a System of Record (Pre-2017)}

The genesis of artificial intelligence as a practical field bears a striking resemblance to the birth of human writing. Both began not with the grand ambition of expressing boundless creativity or philosophical depth, but with the pragmatic need to manage complexity through structured, reliable records. This initial phase of AI, dominated by symbolic reasoning and expert systems, can be understood as its "Cuneiform Moment."

\subsection{The Nature of Early Writing: Concrete and Administrative}

The earliest known writing systems, such as Mesopotamian cuneiform, emerged around 3500-3200 BCE, not for poetry or storytelling, but for administration \cite{walker1987cuneiform}. The growing complexity of urban societies, with their centralized economies, taxation, and trade, created an information management problem that outstripped human memory. Temple and palace officials needed a persistent and unambiguous way to track goods like grain, beer, and livestock. The defining characteristics of these early scripts were their concreteness and domain-specificity. Initially, writing was purely pictographic: a symbol for a bull was a picture of a bull. Over time, these symbols became more abstract and stylized into the characteristic wedge-shaped marks of cuneiform, but their function remained largely tied to recording specific objects, quantities, and transactions. Cuneiform was a system for recording reality according to established conventions, not for generating novel abstract thought. Its power lay in its ability to create a stable, shared source of truth for administrative and legal functions, such as the famous Code of Hammurabi. However, this system was inherently limited. It was labor-intensive, requiring years of training for a specialized class of scribes to master its vast and complex lexicon of symbols. More fundamentally, it was not a generative system for open-ended reasoning. While it could record a law, it could not easily be used to debate the philosophy of justice.

\subsection{The Architecture of Expert Systems: Codified Knowledge}

The first commercially successful wave of AI, from the 1970s through the 1980s, was dominated by expert systems, a paradigm that mirrors the function and limitations of cuneiform with remarkable fidelity. These systems, products of the "Good Old-Fashioned AI" (GOFAI) school \cite{haugeland1985artificial}, were designed to capture the knowledge of human experts in a specific, narrow domain and replicate their decision-making processes. Prominent examples included MYCIN for diagnosing blood infections and eXpert CONfigurer (XCON) for configuring computer systems \cite{buchanan1984rule}. The architecture of a typical expert system consisted of two core components, directly analogous to the elements of a cuneiform-based bureaucracy. The first component, the \textbf{Knowledge Base}, was the heart of the system, a repository of facts and rules about a specific domain, painstakingly elicited from human experts by "knowledge engineers". This knowledge was often represented as IF-THEN rules (e.g., "IF the patient has a fever AND has a rash, THEN consider measles"). This explicit, hand-coded database of domain-specific information is the direct counterpart to the scribe's lexicon of cuneiform symbols, where each rule and fact corresponds to a specific entity or relationship in the world. The knowledge base contained both factual knowledge and heuristic knowledge. The second component, the \textbf{Inference Engine}, was the system's "brain," a set of logical procedures (such as forward chaining or backward chaining) that applied the rules in the knowledge base to the facts of a new situation to derive a conclusion. This mechanism is analogous to the grammatical and procedural rules a scribe would use to combine symbols to form a valid and meaningful record.

The promise of expert systems was immense: to preserve and democratize expertise, making the knowledge of the few available to the many, ensuring consistent and unbiased decision-making, and automating complex cognitive tasks. However, like cuneiform, their fundamental architecture imposed severe limitations that ultimately led to the "AI Winter" of the late 1980s and 1990s. These systems were notoriously brittle; if presented with a situation that fell even slightly outside the scope of their hand-coded rules, they would fail catastrophically \cite{dreyfus1992what}. The process of extracting knowledge from human experts and encoding it into formal rules, the "Knowledge Acquisition Bottleneck," was incredibly slow, expensive, and difficult. Furthermore, these systems were static, unable to learn from new experiences or refine their own knowledge without manual reprogramming. The decline of the expert systems paradigm was not merely a consequence of insufficient funding or processing power, but a fundamental representational crisis. The GOFAI approach, relying on high-level, human-interpretable symbols, hit the same conceptual wall as logographic writing systems. A system whose representations are inextricably tied to concrete objects cannot scale to handle the ambiguity and generativity of abstract thought. The AI Winter was the inevitable consequence of a paradigm that lacked a sufficiently abstract and universal representational layer. For AI to progress toward generality, it first needed to invent its alphabet.

\section{The Alphabet Moment: The Universalization of Representation (2017–2023)}

The transition from pictographic and logographic writing systems to the alphabet was one of the most profound cognitive revolutions in human history. It was not an incremental improvement but a radical shift in the technology of thought, a leap in abstraction that unlocked new potentials for expression, analysis, and knowledge dissemination. In 2017, the field of artificial intelligence experienced its own "Alphabet Moment" with the introduction of the Transformer architecture \cite{vaswani2017attention}, a development that similarly broke from the past and unleashed the generative capabilities that define the modern AI landscape.

\subsection{The Cognitive Revolution of the Alphabet}

Unlike cuneiform, which used thousands of symbols to represent whole words or concepts, the alphabet used a small, finite set of characters to represent the basic sounds of a language (phonemes). This innovation, most notably developed and spread by the Phoenicians, had transformative effects. The core of this revolution was abstraction. The alphabet decoupled the written sign from the object it signified. The letter 'B' does not look like a boat or a bull; it represents an abstract sound. This abstraction meant that any word could be constructed by combining these simple primitives, providing a powerful, efficient, and infinitely generative system for encoding language. This efficiency had a democratizing effect, dramatically lowering the barrier to literacy. Beyond its social impact, the alphabet had deep cognitive consequences. The "alphabet effect" hypothesis posits that the analytic skills required to decode phonetic script promoted new modes of thought, including abstraction, analysis, and classification. This new way of processing information is believed to have created a cognitive environment conducive to the development of codified law, abstract science, and deductive logic \cite{logan2004alphabet}. Neuroscientific studies lend support to this, showing that the brain develops specialized regions, such as the Visual Word Form Area (VWFA), dedicated to the rapid processing of letter forms, a process shaped by motor experiences like writing \cite{dehaene2009reading}.

\subsection{The Transformer as AI's Alphabet}

The publication of the paper "Attention Is All You Need" in 2017 marked the beginning of AI's Alphabet Moment. The Transformer architecture, which was introduced, broke from the sequential processing of its predecessors and proposed a new model based entirely on a few powerful, universally applicable primitives. These primitives function as AI's alphabet, providing a universal system for representing and relating information. The core mechanisms of the Transformer are \textbf{Tokenization}, the process of breaking down any input data into a vocabulary of discrete units called "tokens"; \textbf{Embeddings}, where each token is mapped to a high-dimensional vector representing its meaning in relation to all other tokens; and \textbf{Self-Attention}, the revolutionary "grammar" of the Transformer. For each token in a sequence, the self-attention mechanism dynamically calculates an "attention score" that weighs the relevance of every other token. To do this, each input token embedding is projected into three distinct vectors: a Query ($Q$), a Key ($K$), and a Value ($V$). The attention score is calculated by taking the dot product of the current token's Query with every other token's Key. These scores are scaled and passed through a softmax function to create a probability distribution, which determines how much "attention" to pay to each token. Finally, these attention weights are used to compute a weighted sum of all the Value vectors, producing a new, context-rich representation for the current token. This allows the model to understand complex dependencies within the data.

The power of this architecture lies in its universality and parallelism. Unlike domain-specific expert systems, the Transformer is agnostic to the type of data it processes. The same attention mechanism can find relevant patterns in sentences, pixels, or protein sequences. Furthermore, because attention can be calculated for all tokens simultaneously, the architecture is highly parallelizable, enabling it to scale to unprecedented sizes on modern hardware like Graphics Processing Units (GPUs). This leap to a universal, abstract representation has created models of immense power, but it has also introduced a new and profound challenge: opacity. The very abstraction that gives the Transformer its power also makes its internal workings inscrutable. This "black box" problem is a direct consequence of this evolutionary stage. We do not understand the internal "thoughts" of an LLM for the same reason that the meaning of a word is not contained within the physical shape of its constituent letters. Meaning, as Wittgenstein argued, is determined by its use within a "language game" \cite{wittgenstein1953philosophical}. LLMs learn the language game of human text by modeling the statistical relationships between tokens on a planetary scale. The inscrutability of LLMs is the cognitive price of moving from concrete, interpretable pictographs (the IF-THEN rules of an expert system) to abstract, relational symbols (token embeddings). Having achieved this powerful but opaque form of representation, the next necessary stage in AI's evolution must be the development of tools to bring structure, transparency, and control to these new cognitive capabilities.

\section{The Metalinguistic Moment: AI's Grammatical Turn (The Present)}

Just as the adoption of the alphabet created the conditions for a new kind of self-examination in human thought, the rise of the universal, opaque intelligence of LLMs has forced the AI field into a similar "grammatical turn." This is the Metalinguistic Moment: the stage where intelligence begins to develop tools to analyze, structure, and govern its own cognitive processes. We are living through the dawn of this era, witnessing the emergence of techniques that represent the first, tentative steps toward AI self-reflection.

\subsection{The Birth of Metacognition in Ancient Greece}

The widespread adoption of alphabetic writing in ancient Greece provided a medium for thought itself to become an object of study. Arguments could be written down, scrutinized, and analyzed for their structure, independent of their specific content. This catalyzed a metacognitive revolution, led by philosophers like Plato and, most systematically, Aristotle. They invented the tools to talk about talking, and to reason about reasoning. This revolution had three key pillars. The first was \textbf{Grammar}, where thinkers began to systematically categorize the elements of language. The second was \textbf{Logic} \cite{smith1989logic}, where Aristotle, in his \textit{Organon}, created the first formal system of logic in the West, introducing the syllogism and analyzing its validity based purely on its form. The third was \textbf{Rhetoric}, the formalization of the art of persuasion. Together, these disciplines represented humanity's first major leap into metacognition. They were tools for making the implicit rules of thought explicit, thereby making them transparent, teachable, and improvable.

\subsection{AI's Emerging Self-Reflection}

Today, we are witnessing the emergence of analogous tools in AI, designed to bring structure and transparency to the powerful but opaque reasoning of LLMs. These techniques are the nascent grammar and logic of machine thought. The key technologies defining this Metalinguistic Moment are \textbf{CoT Prompting}, a technique that elicits reasoning from LLMs by prompting them to produce a series of intermediate, step-by-step reasoning steps before giving the final answer. This transforms the model's reasoning from an inaccessible internal state into an explicit, textual object \cite{wei2022chain}. Another key technology is \textbf{Constitutional AI (CAI)}, a framework where an LLM is trained to critique and revise its own outputs based on a set of explicit, human-written principles—a "constitution." It uses a process of supervised self-critique followed by Reinforcement Learning from AI Feedback (RLAIF) to enforce explicit, normative rules on the AI's own linguistic behavior \cite{bai2022constitutional}. Finally, \textbf{Agentic Frameworks (e.g., ReAct)} aim to structure AI reasoning for interaction with the world. The ReAct framework prompts an LLM to operate in an explicit, interleaved loop of Thought, Action (a command to an external tool), and Observation (the result from the tool) \cite{yao2022react}, creating a transparent and auditable cognitive cycle.

These metalinguistic technologies serve a common evolutionary purpose. The "Alphabet Moment" created models with immense generative power but little inherent interpretability or reliability. The tools of the Metalinguistic Moment are the first systematic attempts to build scaffolding around this raw power. CAI is a tool for governance, ReAct is a tool for accountability, and CoT is a tool for interpretability. Together, they represent the beginning of a true "governance architecture" for AI, shifting the focus from simply building a more powerful engine to engineering the safety and control systems that can reliably steer it. Table \ref{tab:metalinguistic} provides a systematic comparison of these foundational metalinguistic frameworks.

\begin{table}[h]
  \caption{A Comparative Analysis of Foundational Metalinguistic AI Frameworks. This table contrasts the primary goals, mechanisms, and key limitations of the technologies defining the current Metalinguistic Moment.}
  \label{tab:metalinguistic}
  \centering
  \begin{tabular}{llll}
    \toprule
    Framework & Primary Goal & Core Mechanism & Key Limitation \\
    \midrule
    CoT \cite{wei2022chain} & Improve reasoning & Step-by-step traces & Error propagation \\
    CAI \cite{bai2022constitutional} & Align behavior & Self-critique \& RLAIF & Constitution quality \\
    ReAct \cite{yao2022react} & Interact reliably & Thought-Action loop & Can get stuck/stray \\
    \bottomrule
  \end{tabular}
\end{table}

\section{The Mathematical Symbolism Moment: The Calculus of Thought (The Next Frontier)}

The development of grammar and logic gave humanity the tools to structure arguments expressed in natural language. However, for problems related to change and continuity, language alone was insufficient. The next great cognitive leap, the invention of calculus in the 17th century, required a new symbolic system that transformed complex reasoning into computable operations \cite{boyer1959history}. As AI moves beyond its Metalinguistic Moment, it faces a similar frontier: the need to develop its own "calculus of thought" to make its reasoning processes not just explicit, but formal, verifiable, and computable.

\subsection{The Leap to Calculus: A New Language for Reasoning}

The invention of calculus by Newton and Leibniz was a monumental achievement. Its true innovation was the creation of a formal operational language for reasoning about change. Calculus introduced a new set of symbols (such as $dy/dx$ for differentiation and $\int$ for integration) and a precise set of rules for their manipulation. This symbolic system made a whole class of complex reasoning computable. Problems that previously required painstaking geometric argumentation could now be solved through the almost mechanical application of algebraic rules. It provided a powerful, abstract notation for describing and calculating the dynamics of the physical world.

\subsection{The Quest for AI's Calculus: From Language to Program}

The next frontier for AI is to make a similar leap, moving from generating reasoning in natural language to generating it within a formal, verifiable, and computable structure. The goal is to create an AI whose "thoughts" are formal objects that can be executed, inspected, and proven correct. This is the "Mathematical Symbolism Moment," and its leading edge is found in the convergence of neuro-symbolic methods and program synthesis. \textbf{Neuro-Symbolic AI} is a promising precursor, seeking to combine the intuitive pattern-matching of neural networks with the rigorous, explainable nature of formal logic \cite{marcus2020next}. A landmark example is DeepMind's AlphaGeometry, which solves Olympiad-level geometry problems. It uses a neural language model as an "intuition engine" to suggest potentially useful auxiliary constructions, and a symbolic deduction engine, armed with the formal rules of Euclidean geometry, to take these suggestions and rigorously explore their logical consequences until a proof is found \cite{trinh2024solving}.

The ultimate expression of this era's goal is \textbf{Program Synthesis}, where an AI's reasoning process manifests as the generation of a computer program. Instead of outputting a natural language explanation, the AI would output a piece of code that, when executed, solves the problem \cite{gulwani2017program}. This makes the reasoning process itself an executable artifact that can be formally verified for correctness, analyzed for efficiency, and safely composed with other programs. This transition is an evolutionary necessity driven by a fundamental flaw in the current paradigm: the compounding error crisis. Long chains of reasoning generated in natural language are prone to amplifying small factual errors or logical missteps, leading to undetectable failure. A "calculus of thought" based on neuro-symbolic methods or program synthesis provides the error-correction mechanism that AI currently lacks. The neural component can propose a plan, but the symbolic or programmatic component can be formally checked at each step. This move is essential for creating reliable and safe autonomous agents capable of performing complex, multi-step tasks in the real world.

\section{The Formal Logic System Moment: Towards a Generative Grammar of Machine Cognition (The Ultimate Goal)}

The historical trajectory of human cognitive tools points toward a final, aspirational frontier: the creation of a perfect, formal system of thought, where reasoning is as clear and verifiable as mathematical calculation. This dream, from Leibniz's \textit{characteristica universalis} to the formal logic of Frege and Russell, represents the pinnacle of structured reasoning. For AI, this translates into the ultimate goal: to evolve from a system that performs intelligent tasks to one whose intelligence is grounded in a formal logic, making it provably beneficial and aligned with human values.

\subsection{The Vision of a Perfect Language and Provable Systems}

The ambition of early 20th-century logicians was to place mathematics on an unshakable foundation of formal axioms and inference rules. While Gödel's incompleteness theorems later showed the inherent limits of any single formal system, the pursuit itself gave birth to modern logic and computer science. For AI, the ultimate goal is not just to create an artificial General Intelligence (AGI) that can match human capabilities, but one that is provably safe and aligned. This requires moving beyond empirical testing to formal guarantees of behavior, the domain of \textbf{Formal Verification}. Applying these techniques to AI is a formidable challenge \cite{amodei2016concrete}. Current approaches include Abstract Interpretation, which soundly overapproximates neural network behavior, and methods based on SMT/MILP solvers, which translate the verification problem into a constraint satisfaction problem. These methods represent the foundational tools for building the trustworthy AI envisioned in this moment \cite{katz2017reluplex}.

\subsection{AGI as a Formal System: A Parallel Cognitive Architecture}

The endpoint of this evolutionary path is an AGI built not as a monolithic, opaque neural network, but as a cognitive architecture grounded in a formal logical system. Such a system would be a distinct, parallel form of intelligence with its own "generative grammar" of cognition—a set of core axioms and inference rules that are explicitly defined, inspectable, and verifiable. This vision addresses the deepest philosophical questions surrounding machine intelligence by focusing on a pragmatic goal: creating a powerful intelligence that is transparent and trustworthy by design. Its value would depend not on subjective experience (qualia) but on whether its operations can be formally proven to be safe, ethical, and aligned. This pursuit will force a convergence between capabilities-driven research and safety-driven research. A central tension exists in the field: the most powerful models (LLMs) are the most opaque, while the most verifiable models (symbolic systems) are the least powerful. The preceding "Mathematical Symbolism Moment" is the first bridge across this divide. The Formal Logic System Moment represents its completion. As AI systems become more deeply integrated into critical societal infrastructure, the demand for formal guarantees will become a legal, commercial, and ethical necessity. The ultimate trajectory of AI development is not just a race for raw intelligence, but a drive toward increasing logical structure and verifiability.

\section{Practical Implications: A Roadmap for Startups in the Euclid's Elements Moment}

This five-stage framework provides a lens for identifying commercial opportunities. By understanding which evolutionary "moment" we are in and which is next, entrepreneurs and investors can strategically position themselves. We refer to the current transitional period between the Metalinguistic and Mathematical Symbolism moments as the "Euclid's Elements Moment"—a time when AI, like Euclid's systematic codification of geometry around 300 BCE, is developing foundational principles and formal structures that will underpin all future reasoning. Just as Euclid's \textit{Elements} provided the axiomatic foundation for mathematical thought for over two millennia, today's AI field is establishing the fundamental "axioms" of machine cognition through techniques like Constitutional AI, formal verification methods, and structured reasoning frameworks. This section translates our framework into three concrete startup ideas leveraging the current Metalinguistic Moment and the emerging Mathematical Symbolism Moment. Table \ref{tab:startups} summarizes these opportunities.

\begin{table}[h]
  \caption{Actionable Startup Opportunities in the Euclid's Elements Moment. This table outlines three business models, each aligned with a specific stage of AI's cognitive evolution, demonstrating the practical application of the paper's theoretical framework.}
  \label{tab:startups}
  \centering
  \begin{tabular}{p{3.2cm}p{2.3cm}p{3.5cm}p{3.0cm}}
    \toprule
    Startup Concept & Era & Business Problem & Target Market \\
    \midrule
    Constitutional Compliance as a Service & Metalinguistic & High cost of ensuring AI content meets regulations & Finance, Healthcare, Legal Tech \\\\
    Neuro-Symbolic Robotics for Logistics & Mathematical Symbolism & Brittleness and poor error recovery in warehouse robots & E-commerce, 3PL, Manufacturing \\\\
    Generative Program Synthesis & Mathematical Symbolism & Slow, expensive custom automation of enterprise workflows & Enterprise Information Technology (IT), BPO, No-code users \\
    \bottomrule
  \end{tabular}
\end{table}

\subsection{Example 1: Constitutional Compliance as a Service (Metalinguistic Era)}

Enterprises in regulated industries (finance, healthcare) face significant risk in deploying LLMs due to potential violations of regulations like the Health Insurance Portability and Accountability Act (HIPAA) or the Financial Industry Regulatory Authority (FINRA). A startup could offer "Constitutional Compliance as a Service," delivering fine-tuned models with pre-built, auditable "constitutions" tailored to specific regulatory landscapes. Using the Constitutional AI (CAI) framework \cite{bai2022constitutional}, the startup would codify regulations into machine-readable principles, train a "critic" model to self-critique outputs against this constitution, and fine-tune a production model using RLAIF to generate inherently compliant content. A key feature would be an audit trail explaining why a response was revised, providing a defensible record for regulators. The target market would be compliance and legal departments, offering a scalable solution to move compliance from a manual review process to an automated, built-in feature of the AI itself.

\subsection{Example 2: Neuro-Symbolic AI for Autonomous Robotics (Mathematical Symbolism Era)}

Warehouse automation is booming, but current robots are brittle and struggle with dynamic environments. A startup could build a robotics intelligence platform based on a neuro-symbolic agent architecture. This system would feature a Vision-Language Model (VLM) "Brain" for high-level, commonsense planning (e.g., decomposing an order into steps) and a symbolic "Spinal Cord" for verifiable action execution using classical Task and Motion Planning (TAMP). The key innovation is an error recovery loop based on the ReAct framework. If a symbolic action fails (e.g., a grasp), the symbolic engine generates a formal observation (ERROR: Grip\_failed). This is fed back to the VLM "Brain," which reasons about the failure and generates a new, corrective plan ("Try a different grasp strategy") \cite{yao2022react}. This closed-loop feedback mechanism would dramatically increase operational uptime and adaptability, targeting large e-commerce fulfillment centers and logistics providers.

\subsection{Example 3: Generative Program Synthesis for Enterprise Automation (Mathematical Symbolism Era)}

Automating cross-application business processes in large enterprises is a slow, expensive process requiring skilled developers. A startup could create an AI platform for generative enterprise automation \cite{gulwani2017program}. A non-technical user could describe a workflow in structured natural language (e.g., "When a candidate is 'Hired' in Greenhouse, create a user in Okta and send a contract via DocuSign"). The AI agent would synthesize executable code that makes the correct Application Programming Interface (API) calls. Crucially, a built-in formal verification module would analyze the generated program to prove properties like "A contract is never sent unless the user is created in Okta," ensuring the automation is not just functional but reliable and correct-by-construction. This targets enterprise IT departments, combining the accessibility of natural language with the reliability of formally verified code.

\section{Closing the Loop: How the Geometry of Cognition Reshapes AI's Foundations}

The evolutionary framework presented thus far treats the development of AI as a process that parallels the history of human cognitive technologies. However, this analogy is not a one-way street. As AI advances through these stages, the very tools and concepts it develops will create a feedback loop, fundamentally reshaping its own underlying architecture. The future of AI will not just be about climbing this ladder of cognitive abstraction, but about using the insights gained at each rung to rebuild the ladder itself. This section explores how this reflexive, self-modifying dynamic might manifest, particularly in the foundational layers of representation and reasoning.

The "Alphabet Moment," for example, has been dominated by a single paradigm for representation: tokenization, which is conceptually analogous to the Latin alphabet. Just as English words are composed from a small set of letters, LLMs process information by breaking it down into a finite vocabulary of sub-word units. This approach has been incredibly successful, granting models their universality. However, the history of human language offers alternative architectural blueprints. Consider the compositional nature of Chinese characters, where semantic radicals (\textit{e.g.}, a water radical) combine with phonetic components to create a vast lexicon of concepts in a structured, meaningful way. A future LLM architecture might abandon the purely linear, statistical nature of current tokenization for a compositional, graph-like system of representation. In such a model, a new concept would not be represented by a new, arbitrary "word" (token sequence) but by a novel combination of existing semantic and relational primitives. This would be a move from an alphabetic to a logographic-compositional substrate. This could offer significant advantages in zero-shot or few-shot learning, as the model could reason about the potential meaning of a new concept based on its constituent parts, much as a human can guess the meaning of a word like "hydro-electric" by understanding "hydro" and "electric." It would represent a foundational shift from processing sequences of tokens to navigating a semantic graph of concepts, potentially leading to more robust and interpretable models.

Similarly, the "Metalinguistic Moment" will have profound implications for the internal language of thought within AI. Currently, techniques like CoT prompt the model to externalize its reasoning process into human-readable natural language. This is akin to a student showing their work. While valuable for interpretability, it forces the AI to "think" in a medium that is inherently ambiguous and inefficient for formal reasoning. The feedback loop suggests that future architectures will internalize this process. AIs will develop their own internal metalanguages—optimized, compressed, and formally precise—for self-analysis and planning. Instead of generating a verbose English paragraph to structure its thoughts, an advanced AI might generate a concise string in a self-developed formal grammar. This internal monologue, while perhaps inscrutable to humans directly, could be provably mapped back to human-understandable concepts when required for auditing. This allows the model to escape the cognitive friction of natural language for its internal operations, potentially leading to vast improvements in reasoning speed and logical consistency.

Finally, the "Mathematical Symbolism Moment" will not remain a layer applied on top of a neural substrate, as seen in current neuro-symbolic hybrids. The feedback loop will drive these symbolic structures deeper into the core architecture. Future generations of hardware and models may not be based on homogeneous layers of neurons but on heterogeneous computational fabrics that natively incorporate verifiable logical and mathematical operators. Imagine a "transformer" whose "attention" mechanism is not just a statistical weighting but can be instantiated as a formally defined operation—a set-theoretic intersection, a logical conjunction, or a differentiable function chosen from a library of verifiable components. This would represent the ultimate fulfillment of the paper's thesis: the evolution would come full circle, embedding the "calculus of thought" not as an application running on the AI, but as the fundamental physics of the AI itself. This closes the loop, creating a system where the architecture for reasoning and the act of reasoning are one and the same, paving the way for the provably aligned and reliable systems envisioned in the Formal Logic System Moment.

\section{Conclusion: The Inevitability of Structure}

This paper has charted a course through the evolutionary history and future trajectory of AI, framed by a "Geometry of Cognition." We have argued that AI's development is a structured journey through five distinct moments mirroring humanity's own cognitive ascent. From the \textbf{Cuneiform Moment} of brittle expert systems, AI progressed to the \textbf{Alphabet Moment} with the Transformer, which unleashed generative capability at the cost of opacity. This has propelled us into the present \textbf{Metalinguistic Moment}, an era defined by the need to manage these opaque models with nascent tools of governance and interpretability like CoT and CAI.

Looking forward, the path leads to the \textbf{Mathematical Symbolism Moment}, where a computable "calculus of thought" will emerge from neuro-symbolic architectures and program synthesis. This stage is a necessary solution to the compounding error crisis limiting today's agents. The ultimate goal is the \textbf{Formal Logic System Moment}, where the pursuit of capability converges with the pursuit of safety, culminating in an AGI whose intelligence is grounded in a formal system, making it provably aligned and trustworthy.

However, our analysis reveals that this evolutionary process is not merely sequential but fundamentally reflexive. The "Closing the Loop" dynamic demonstrates that as AI advances through these cognitive stages, the tools and insights it develops create a powerful feedback mechanism that recursively transforms its own foundational architecture. The metalinguistic capabilities emerging today will not simply layer atop existing neural substrates—they will drive the development of entirely new representational paradigms, from compositional semantic graphs to native symbolic-neural hybrid architectures. This reflexive quality suggests that the path to AGI is not just about climbing a ladder of cognitive sophistication, but about using each rung to rebuild the ladder itself in more powerful and principled ways.

The overarching conclusion is the inevitability of structure coupled with the transformative power of cognitive recursion. As intelligence grows in power, it must develop more sophisticated layers of formal structure to manage itself and ensure reliability. Yet this structuring process becomes increasingly self-directed, with each cognitive breakthrough enabling more fundamental architectural innovations. The path to AGI is thus both a drive toward greater abstraction, self-reflection, and logical rigor, and a process of continuous self-reconstruction guided by these emerging principles.


\bibliographystyle{plain}
\bibliography{sample-base}

\end{document}